\documentclass[runningheads,a4paper]{llncs}

\usepackage[colorlinks,
            linkcolor=black,
            anchorcolor=black,
            citecolor=black
            ]{hyperref}

\usepackage{amssymb}
\usepackage{times}
\usepackage{epsfig}
\usepackage{graphicx}
\usepackage{amsmath}
\usepackage{amssymb}
\usepackage{multirow}
\usepackage[linesnumbered,ruled]{algorithm2e}
\usepackage{caption}
\usepackage{subcaption}
\usepackage{epstopdf}
\usepackage[flushleft]{threeparttable}
\usepackage{float}
\usepackage{adjustbox}

\begin{document}
\mainmatter  

\title{Deep Representation Learning For Multimodal Brain Networks}
\titlerunning{Deep Multimodal Brain Networks}

%
%
\author{Wen Zhang$^1$, Liang Zhan$^2$, Paul Thompson$^3$, Yalin Wang$^1$}
\authorrunning{Zhang et al.}

\institute{$^1$School of Computing, Informatics and Decision Systems Engineering, \\Arizona State University, AZ, USA\\
$^2$Electrical and Computer Engineering, University of Pittsburgh, PA, USA\\
$^3$Imaging Genetics Center, University of Southern California, CA, USA}

\toctitle{}
\maketitle

\begin{abstract}
Applying network science approaches to investigate the functions and anatomy of the human brain is prevalent in modern medical imaging analysis. Due to the complex network topology, for an individual brain, mining a discriminative network representation from the multimodal brain networks is non-trivial. The recent success of deep learning techniques on graph-structured data suggests a new way to model the non-linear cross-modality relationship. However, current deep brain network methods either ignore the intrinsic graph topology or require a network basis shared within a group. To address these challenges, we propose a novel end-to-end deep graph representation learning (Deep Multimodal Brain Networks - DMBN) to fuse multimodal brain networks. Specifically, we decipher the cross-modality relationship through a graph encoding and decoding process. The higher-order network mappings from brain structural networks to functional networks are learned in the node domain. The learned network representation is a set of node features that are informative to induce brain saliency maps in a supervised manner.  We test our framework in both synthetic and real image data. The experimental results show the superiority of the proposed method over some other state-of-the-art deep brain network models. 
\end{abstract}

\section{Introduction}
There is growing scientific interest in understanding functional and structural organizations of the human brain from a large scale of multimodal brain imaging data. In medical imaging analysis, one of the popular ways for this task is to explore brain regional connections (i.e., brain networks) measured from the brain imaging signals. The topological patterns of brain networks are closely related to the brain functional organizations~\cite{bullmore2012economy} and the connection breakdown between the relevant brain regions has an intimate association with the progress of neurodegenerative diseases~\cite{jao2015functional,repovs2011brain} or normal brain developments~\cite{zhang2018multimodal}. However, patterns of focal damages in brain networks are different across modalities, making the mining of multimodal network changes difficult.

Deep learning methods have been successfully applied to extract biological information from the neuroimaging data~\cite{ronneberger2015u,suk2013deep}. Most of the prior brain network analysis represent graph structure as a grid-like image to enable convolutional computation~\cite{plis2018reading,deshpande2015fully,wang2017structural}. 
More recently, deep graph convolutional networks (GCNs) have been introduced to brain network research~\cite{arslan2018graph,kawahara2017brainnetcnn,ktena2018metric}. These studies perform the localized convolutional operation at either graph nodes or edges. They can be categorized into the graph spectral convolution~\cite{arslan2018graph,ktena2018metric} and the graph spatial convolution~\cite{hamilton2017inductive}. The former approach is suitable for node-centric problems defined on the fixed-sized neighborhood graphs. For graph-centric problems, the spectral method requires a group-wise graph structure before approximating the spectral graph convolution. Therefore, its performance to a large extent depends on the predefined network basis.  However, the existing framework~\cite{kawahara2017brainnetcnn} is designed for a single modality and lacks a well defined k-hop convolutional operator on each node. This makes the multimodal brain network fusion intractable in the node domain and thus difficult to draw brain saliency maps.

In this paper, we propose a novel GCN model for multimodal brain networks analysis. Two naturally coherent brain network modalities, i.e., functional and structural brain networks, are considered. The structural network acts as the anatomical skeleton to constrain brain functional activities and, in return, consistent functional activities reshape the structural network in the long term~\cite{bullmore2012economy}. Hence, we argue the existence of a high-level dependency, namely networks communication~\cite{avena2018communication}, across them. It is deciphered by a deep encoding-decoding graph network in our model. Meanwhile, the obtained node features help representation learning of brain network structure in a supervised manner. The contributions can be summarised into four-folds. (1) It is the first paper using a deep graph learning to model brain functions evolving from its structural basis. (2) We propose an end-to-end automatic brain network representation framework based on the intrinsic graph topology.
(3) We model the cross-modality relationship through a deep graph encoding-decoding process based on the proposed multi-stage graph convolutional kernel. (4) We draw graph saliency maps subject to the supervised tasks, enabling phenotypic and disease-related biomarker detection. 
\section{Methodology}
\textbf{Multimodal Brain Network Data.} A brain network uses a graph structure to describe interconnections between brain regions and is a weighted graph $G=\{V,E,X\}$, where $V=\{v_i\}^N_{i=1}$ is the node set indicating brain regions, $E=\{\epsilon_{i,j}\}$ is the edges set and $X=\{x_{i,j}\}$ is the corresponding edge weight; For a given subject, we have a pair of networks $\{G^f,G^d\}$, where $G^f=\{V,E^f,X^f\}$ represents the functional brain network and $G^d=\{V,E^d,X^d\}$ is the structural brain network. These two networks share the same set of nodes, i.e., using an identical definition of brain regions, but differ in network topology and edge weights. An edge weight $x_{i,j}^f$ in $G^f$ is the correlation of fMRI signals between node $v_i$ and $v_j$, while a structural edge weight $x_{i,j}^d$ in $G^d$ is the probability of fiber tractography between them. 

\vspace{-1em}
\subsection{Multi-Stage Graph Convolution Kernel}
\vspace{-0.5em}
A brain structural network can be interpreted as a freeway net where biological information such as brain functional signals flows from node to node. In the brain network, a node shall be affected by its neighboring nodes and their affection is negatively correlated with the shortest network distance~\cite{stam2016relation}. To encode these node-to-node patterns, we adopt the spatial graph convolution kernel which will give the node embedding features with respect to the local graph topology. It defines a way to aggregate node features in a given size of neighborhood, e.g., 1-hop connections.       

Given a target node $v_i$ and its neighbourhood graph topology $G_{\mathcal{N}(v_i)}$, the graph convolution kernel first collects node features $h_{v_i}$ of its immediate neighbours: 
\vspace{-0.5em}
\begin{equation}\label{Eq:AGG}
\vspace{-0.5em}
    AGG(h_{v_i})=\sum_{v_j\in \mathcal{N}(v_i)}h_{v_j}\cdot x_{i,j},
\end{equation}
and then updates the node feature as:
\vspace{-0.1em}
\begin{equation}\label{Eq:update}
    h'_{v_i}=\sigma(AGG(h_{v_i})\cdot w).
\vspace{-0.5em}
\end{equation}
Here, $\sigma$ is a non-linear activation and $w\in\mathbb{R}^{F\times F'}$ is a learnable weight matrix of a fully-connected layer (FC). Previous research proves that a $k$-hop convolution kernel can be divided into $k$ 1-hop convolutions~\cite{kipf2016semi}. Therefore, we stack several 1-hop convolutions to increase size of the effective receptive field on graphs. 

A potential problem with Eq.~\ref{Eq:AGG} is its poor generalization of the local aggregation, i.e., the aggregation weight is fixed to be $x_{i,j}$. Though these predefined values reflect the brain biological profiles, they might not be optimal for brain network encoding, especially for the cross-modality learning pursued by our research. For example, brain regions that are interconnected with large weights in the brain structural network are not guaranteed to be more strongly connected in the brain functional network as well~\cite{osmanliouglu2019system}. Besides, compared with brain structural networks, brain functional networks are more dynamic and fluctuant on the edge connections. Therefore, the dynamic adjustment of the aggregation weights during graph learning is favored. To this end, we adopt the idea of graph attention network (GAT)~\cite{velickovic2017graph}. Given each pair of node features, their dynamic edge weights are learned by a single-layer feedforward neural network, i.e.,  $X^{ATT}=\{x^{ATT}_{i,j}\}=\{f_{att}(h_{v_i},h_{v_j})\}$. More specifically, we first increase the expression power of the node features by using a shared linear transformation, $\Tilde{h}_{v_i}=h_{v_i}\cdot w$, where $w\in \mathbb{R}^{F\times F'}$ is a learned parameter. Then, we use a single-layer feedforward neural network to derive the edge weight:
\begin{equation}\label{Eq:feature}
\vspace{-0.5em}
    \Tilde{x}_{i,j}=\sigma(a^T[\Tilde{h}_{v_i}\oplus \Tilde{h}_{v_j}]),
\end{equation}
where $\oplus$ is the concatenate operator and $a\in \mathbb{R}^{2F'}$ is a parameter of the feedforward network. To assure generalization of Eq.~\ref{Eq:feature} across different nodes, a softmax layer is append for normalization of the neighbourhood,
 \begin{equation}
 \vspace{-0.8em}
     x^{ATT}_{i,j}=\frac{exp(\sigma(a^T[\Tilde{h}_{v_i}\oplus \Tilde{h}_{v_j}]))}{\sum_{k\in\mathcal{N}(v_i)}exp(\sigma(a^T[\Tilde{h}_{v_i}\oplus \Tilde{h}_{v_k}]))}.
 \end{equation}

\begin{figure}[t]
\vspace{-1em}
\center
        \includegraphics[width=0.95\linewidth]{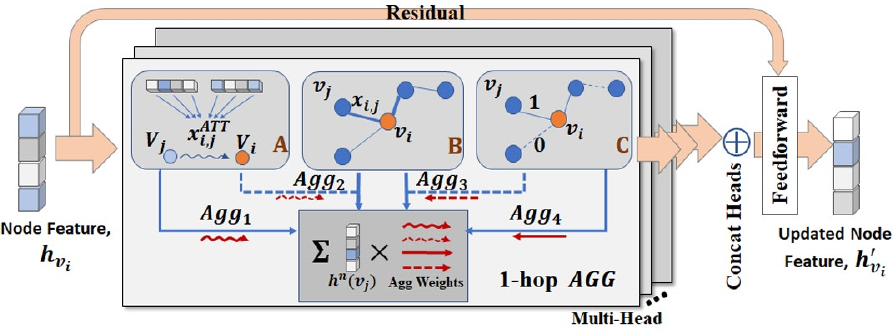}
    \caption{Multi-stage graph convolution kernel (MGCK). Three aggregation mechanisms are dynamical combined, including the graph attention weight $x^{ATT}_{i,j}$ (A), the original edge weight $x_{i,j}$ (B), and the binary weight $\delta(x_{i,j})$ (C).}
    \label{fig:MGCK}
\vspace{-1em}
\end{figure}

Compared with $x_{i,j}$, $x^{ATT}_{i,j}$ is associated with the node order and thus is asymmetric on edge $\epsilon_{i,j}$. Besides, it is free of local network topology. In addition to the graph attention based aggregation (Fig.~\ref{fig:MGCK}, A), we also propose a binary symmetric aggregation defined with a threshould function $\delta(x_{i,j})$ (Fig.~\ref{fig:MGCK}, B). $\delta(x_{i,j})$ thresholds an edge by a given threshould value $\gamma$, e.g., aggregation weight will be 1 if $x_{i,j}>\gamma$, otherwise 0. We set $\gamma=0$ empirically in this study. This process follows an assumption that two brain regions are highly interactive in functional brain network as long as they are structurally connected~\cite{stam2016relation}. To integrate all of the aggregation mechanisms, we design a multi-stage graph convolution kernel (MGCK).  Eq.~\ref{Eq:AGG} is thus updated as:
\begin{equation}\label{Eq:AGG_mod}
\vspace{-0.5em}
\begin{split}
    AGG(h_{v_i})=&\sum_{v_j\in \mathcal{N}(v_i)}h_{v_j}\cdot (x_{i,j}+\alpha)\cdot(x^{ATT}_{i,j}+\beta \delta(x_{i,j}))\\
    =&\sum_{v_j\in \mathcal{N}(v_i)}h_{v_j}\cdot (x_{i,j}x^{ATT}_{i,j}+\beta x_{i,j}+\alpha x^{ATT}_{i,j}+\alpha\beta \delta),
\end{split}
\end{equation}
where $\alpha$ and $\beta$ are learnable parameters balancing different aggregation mechanisms. In the above equation, we have 4 different aggregation weights. $x_{i,j}x^{ATT}_{i,j}$ and $x_{i,j}$ are the pre-defined network connections with and without attention weights. $x^{ATT}_{i,j}$ is the attention aggregation alone and $\delta$ is the threshold connections. In the end, we introduce the multi-head learning~\cite{vaswani2017attention} to stabilize the aggregation in MGCK. $K$ independent multi-stage aggregation are conducted and aggregated features are concatenated before feeding to a FC layer. Accordingly, Eq.~\ref{Eq:update} is updated as:
\begin{equation}\label{Eq:multihead}
\vspace{-0.3em}
    \hat{h}_{v_i}=\oplus^K_{k=1}[\sigma(AGG^k(h_{v_i})\cdot w)].
\end{equation}
Previous research indicates that graph convolution network performs poorly with a deep architecture due to the high complexity of back-propagation in the deep layers. To address this problem, residual block in GCN~\cite{li2019deepgcns} is proposed. It is inspired by the success of ResNet~\cite{he2016deep} for image data. We add the residual connection after MGCK,
\begin{equation}\label{Eq:residule}
\vspace{-0.5em}
    h'_{v_i}=\mathcal{F}(\hat{h}_{v_i},\hat{w})+w_m h_{v_i}.
\end{equation}
$\mathcal{F}$ is a FC layer parameterized by $\hat{w}$. Parameter $w_m$ is designed to match the dimensions.

\begin{figure*}[t]
\vspace{-0.5em}
\center
        \includegraphics[width=1\linewidth]{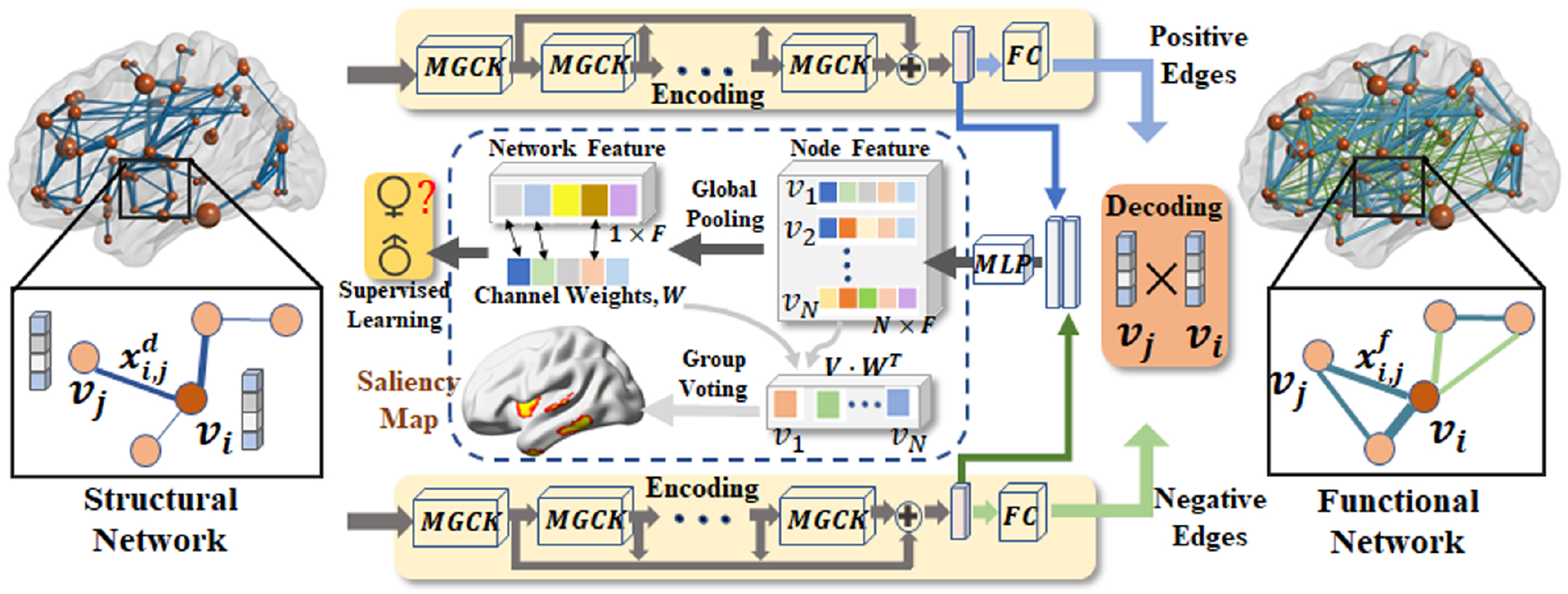}
    \caption{Pipeline of DMBN. The structural network is fed into two independent encoding-decoding networks to generate the cross-modality encoding of the positive and negative functional connections. Meanwhile, the node features from these two networks are combined and serve as the multimodal graph representations for the supervised learning tasks via a MLP network. During this process, a brain saliency map is derived.}
    \label{fig:pipeline}
\vspace{-1.2em}
\end{figure*}

\vspace{-1em}
\subsection{Deep Multimodal Brain Networks (DMBN)}
\vspace{-0.5em}
We show the pipeline of DMBN in Fig.~\ref{fig:pipeline}. It generates the multimodal graph node representations for different learning tasks. There are two parts in DMBN. The first part is for cross-modality learning via an encoding-decoding network. Here, we construct brain functional network from brain structural network. The brain functional network contains both positive and negative connections. These two types of brain functional connectivities yield a distinct relationship with brain structural network~\cite{honey2009predicting,schwarz2011negative}. Hence, we separate their encoding into two independent encoding networks. For each graph encoder, we use several MGCK layers to aggregate node features from diverse ranges of the neighborhood in structural network. The generated node features are then fed into the decoding networks to reconstruct the positive and negative connections respectively.
Specifically, for each undirected edge $e_{i,j}$, we define the reconstructed links as:
\begin{equation}\label{eq:embedding}
\vspace{-0.8em}
    \hat{x}_{i,j}=\frac{1}{1+exp(-h_{v_i}^T\cdot \Theta\cdot h_{v_j})},
\end{equation}
where $h_{v_i}$ is a node feature vector in the network embedding space and $\Theta$ is a learnable layer weight. Eq.~\ref{eq:embedding} maps the deep node embeddings $\{h_{v_i}\}$ to a connection matrix $\{\hat{x}_{i,j}\}$ where each element ranges from 0 to 1 consisting with the functional connections.

The second part of our model is a supervised learning. The node embedding features ($h_v$) from the positive and negative encoding networks are concatenated node-wisely and processed by an MLP. Since our tasks are graph level learning, a global pooling is applied before the last FC layer to remove the effect of node orders. Along with the supervised learning tasks, it is important to understand the key brain regions closely associated with the tasks. Inspired by the classic activation maps~\cite{arslan2018graph}, a graph localization strategy is carried out by learning contribution scores of graph nodes. As shown in Fig.~\ref{fig:pipeline}, suppose the final node feature matrix consists of $F$ channels for $N$ nodes, a global mean pooling generates a channel-wise vector treated as the network feature. Therefore, each channel has a corresponding weight, $w_i$, learned by the last FC layer. To obtain the node-wise importance score, we warp it back by an inner product between node features and channel weights, i.e., $h_v\cdot W^T$. In the end, we rank the top-$k$ nodes for each subject and conduct a group voting to obtain the group-wise saliency map.

There are 3 loss terms in DMBN controlling the brain network reconstruction and supervised learning tasks (Eq.~\ref{Eq:TotalLoss}). The reconstruction loss consists of the global and local decoding losses to preserve different levels of graph topology. 
\begin{equation}\label{Eq:TotalLoss}
\vspace{-0.5em}
    L_{all}=\mu_1 L_{global}+\mu_2 L_{local}+L_{preds},
\vspace{-0.1em}
\end{equation}

\vspace{-1em}
\paragraph{\textbf{1) Global Decoding Loss.}} This term evaluates the averaged performance of edge reconstruction in the target network.
\begin{equation}
\vspace{-1em}
\begin{split}
    \mathcal{L}_{global}=&\frac{1}{|E|}\sum_{e_{i,j}} a_{i,j}(\hat{x}^{f+}_{i,j}-\hat{x}^{f-}_{i,j}-x^f_{i,j})^2,
\end{split}
\end{equation}
where $a_{i,j}$ is the additional penalty of the edge reconstruction. Here, we set it as $e^{abs(x^f_{i,j})}$, which gives the higher weights for stronger connections in brain functional network. $\hat{x}^{f+}$ and $\hat{x}^{f-}$ indicate the decoded network connections from the positive and negative flow of encoding.

\vspace{-1em}
\paragraph{\textbf{2) Local Decoding Loss.}} The cross-modality reconstruction of brain networks is challenging, hence we do not expect a full recovery of all edges but rather the reconstruction of local graph structure on important connections, e.g., edges with strong connections in both structural and functional networks. We adopt the first-order proximity~\cite{wang2016structural} to capture the local structure. 
The loss function is defined as:
\begin{equation}\label{eq:localloss}
\vspace{-1em}
\begin{split}
    \mathcal{L}_{local}=\sum_{i=1}^n \frac{1}{|\mathcal{N}^d_i|}\sum_{j\in\mathcal{N}^d_i}e^{\delta(x^d_{i,j})}||h^{f}_{v_i}-h^{f}_{v_j}||^2_2 ,
\end{split}
\end{equation}
where $|\mathcal{N}^d_i|$ is the number of neighbouring nodes of $v_i$ in brain structural network. $\delta(x^d_{i,j})$ is a threshold function which favors strong generalization.
Eq.~\ref{eq:localloss} generalizes Laplacian Eigenmaps~\cite{belkin2003laplacian} and drives nodes with similar embedding features together. 

\vspace{-1em}
\paragraph{\textbf{3) Supervised Loss.}} The loss function for prediction is defined as:
\begin{equation}\label{Eq:PredLoss}
\vspace{-0.8em}
\begin{split}
    L_{pred}=-\frac{1}{K}\sum_{i=1}^K y_i\cdot log(f_{pred}(h_{v_i})), 
\end{split}
\end{equation}
where $K$ is the number of subjects and $f_{pred}$ is a function learned by the MLP network.



\section{Experiment}

\subsection{Gender Prediction}\label{sec:genderprediction}
\vspace{-0.5em}
\paragraph{\textbf{Dataset.}} The data are from the WU-Minn HCP 1200 Subjects Data Release~\cite{van2013wu}. We include 746 healthy subjects (339 males, 407 females), each has high-quality resting fMRI and dMRI data. The functional network is processed using CONN toolbox~\cite{whitfield2012conn} and structural connectivity is measured by using FSL toolbox~\cite{jenkinson2012fsl}. Here we try to predict the gender based on the multimodal brain network topology. Previous research has shown the strong relationship between gender and brain connectivity patterns~\cite{ruigrok2014meta}.


\vspace{-1em}
\paragraph{\textbf{Experiment Setup.}} We select 5 state-of-the-art baseline models for comparison, where 3 of them, i.e. tBNE~\cite{cao2017tbne}, MK-SVM~\cite{dyrba2015multimodal} and mCCA + ICA~\cite{sui2011discriminating}, are transitional machine learning algorithms while the rest two, i.e. BrainNetCNN~\cite{kawahara2017brainnetcnn} and Brain-Cheby~\cite{ktena2018metric} use deep models. In addition, 5 variant models of MDBN are tested in the experiments as an ablation study. We apply the 5-fold cross-validation for all methods. In our model setting, the positive connection encoding has 5 cascade MGCK layers and negative connection encoding has 4 MGCK layers. In each encoding, each of MGCKs has the feature dimension [128] and 4-heads learning. We report the statistical results with three evaluation metrics: accuracy, precision, and F1 scores. Besides, we take a grid search to decide hyperparameters $\mu_1$ and $\mu_2$. Based on the empirical knowledge, we set the search range for $\mu_1$ as [10, 1, 0.1, 0.01] and $\mu_2$ as [5, 1, 0.5, 0.1]. The best result appears at $\mu_1=1$ and $\mu_2=0.5$. Details can be found in Supplementary Fig.1.

\begin{table}[t]
\vspace{-1em}
\centering
\scriptsize
\caption{Performance of gender prediction in the HCP data.}
\begin{adjustbox}{width=.8\textwidth,center}
\begin{tabular}{c|ccc|ccc}
 \multirow{2}{*}{Method} & \multicolumn{3}{c|}{HCP (Gender)}   & \multicolumn{3}{c}{PPMI (Disease)} \\ \cline{2-7}
   &$Acc$  &$Prec$  &$F1-Score$ &$Acc$  &$Prec$  &$F1-Score$\\ \hline
 
 tBNE~\cite{cao2017tbne}    &0.543  &0.497  &0.503 &0.580  &0.597  &0.530 \\
 MK-SVM~\cite{dyrba2015multimodal}    &0.481  &0.438  &0.524  &0.587  &0.487  &0.568 \\
 mCCA + ICA~\cite{sui2011discriminating}   &0.680  &0.703  &0.691 &0.640  &0.660  &0.622\\
 Brain-Cheby~\cite{ktena2018metric}    &0.739  &0.740  &0.739  &0.635  &0.622  &0.628\\
 BrainNetCNN~\cite{kawahara2017brainnetcnn}    &0.734  &0.775  &0.684  &0.673  &0.695  &0.778\\ \hline
 w/o Recon$^+$ &0.738  &0.692  &0.767  &0.688  &0.727  &0.786\\ 
 w/o TAGG\&Recon$^+$ &0.699  &0.696  &0.738  &- &- &-\\
 w/o AAGG\&Recon$^+$ &0.681  &0.689  &0.735  &- &- &- \\\hline
 MDBN w/o Global &0.784  &0.798  &0.799  &- &- &- \\
 MDBN w/o Local &0.793  &0.814  &0.824  &- &- &- \\
 MDBN           &\textbf{0.819}*  &\textbf{0.836}*  &\textbf{0.845}* &\textbf{0.728}*  &\textbf{0.859}*  &\textbf{0.735} \\\hline
\end{tabular}%
\end{adjustbox}
\label{tab:genderprediction}
\begin{tablenotes}
\small\item $\mathbf{^*}$ stands for significance.$\mathbf{^+}$ indicates the variant model using a single modality.
\end{tablenotes}
\vspace{-1em}
\end{table}
\vspace{-1em}

\vspace{-0.2em}
\paragraph{\textbf{Results.}} As shown in the Tab.~\ref{tab:genderprediction} (HCP), our model achieves the highest accuracy ($ACC>81.9\%$) in the gender prediction among all the methods and significantly outperforms the others with at least 8\% and 10\% increases in accuracy and F1 scores, respectively. Generally, deep models are superior to the traditional node embedding method (tBNE). We notice that, when we remove the cross-modality learning, i.e., variant methods denoted by w/o Recon, the performance drops significantly. Though they are still comparable to the other baselines, the training process is unstable with a high variance. The cross-modality learning enables node-level learning to be effective and consequently affects further graph-level learning. In addition, the 10 most important brain regions affecting the gender prediction are shown in Supplementary Fig. 2. These regions spread at the cortical areas including the frontal and orbital gyrus, precentral gyrus, insular gyrus, as well as the subcortical areas such as basal ganglia. All those regions play vital roles in regulating cognitive functioning, motor and emotion controls, which, with a high probability, exert the gender discrepancy~\cite{rijpkema2012normal,ruigrok2014meta}.


\vspace{-1em}
\paragraph{\textbf{Ablation Analysis.}} We explore influence of each element in our model (Tab.~\ref{tab:genderprediction}). We first remove the decoding network that makes our model a single modality learning (w/o Recon). Under such a configuration, our model is still comparable to the baselines. However, the decreased performance suggests the cross-modality is indispensable to an informative network representation. Based on this setting, we further evaluate the role of each aggregation mechanism in MGCK. We remove the threshold aggregation weight (w/o TAGG$\&$Recon) and graph attention aggregation (w/o AAGG$\&$Recon) respectively. All of them cause a significant decrease in performance. In addition to the single modality learning, we also validate the importance of different reconstruction losses in multimodal learning. Missing the local (MDBN w/o Local) or global (MDBN w/o Global) losses results in around 3\% downgrade in prediction accuracy. Meanwhile, the global reconstruction loss yields a larger weight than the local reconstruction loss. Since the global loss considers all of the edges in the functional network, it contains relatively more fruitful information than the local loss which focuses on the direct edges in the structural network. However, they are complementary to each other.

\begin{figure*}[!t]
\vspace{-1em}
\center
        \includegraphics[width=1\linewidth]{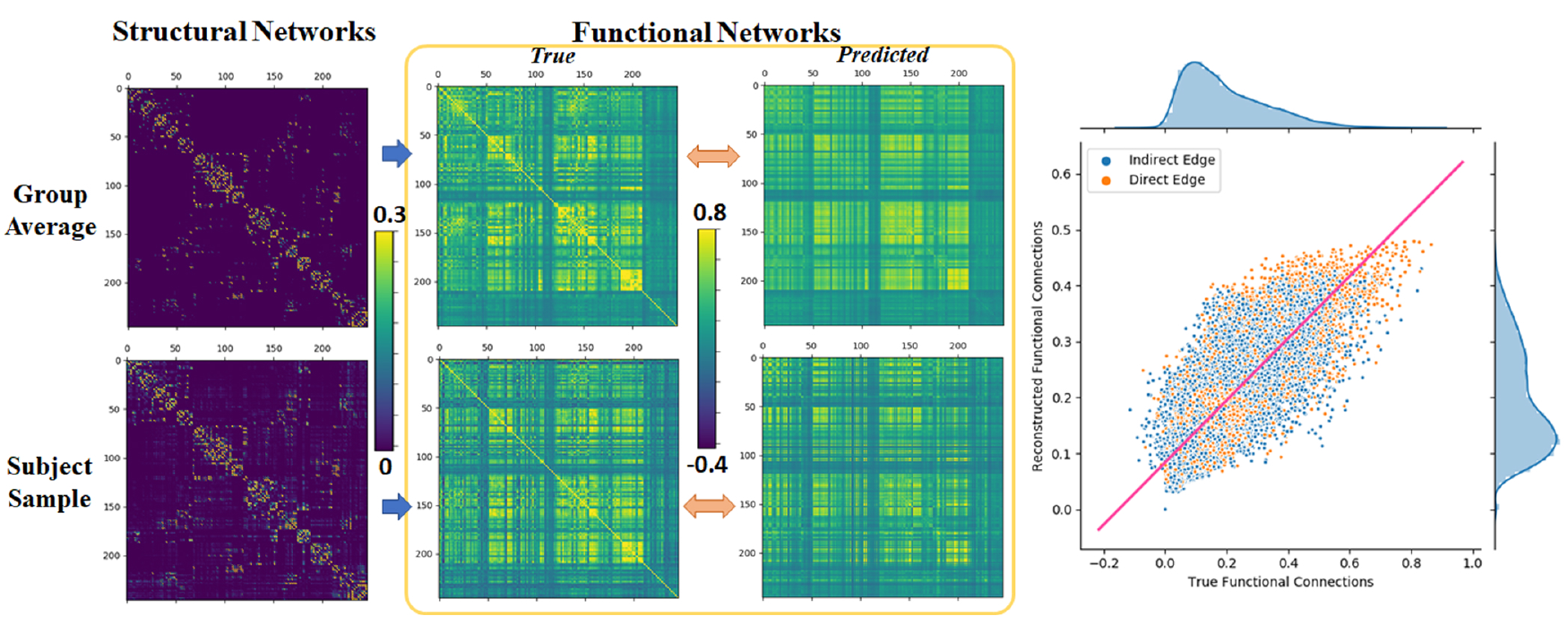}
    \caption{The cross-modality learning results. The functional network is predicted (middle) from its structural network counterpart (left). We present the group averaged result and an individual sample. The statistical evaluation (Spearman correlation, $r_S$) of reconstructed functional networks is conducted (right) and the predicted edge weights are significantly correlated with the ground truth data, $r_s=0.83$.}
    \label{fig:AE_gender}
\vspace{-1em}
\end{figure*}

\vspace{-1em}
\paragraph{\textbf{Cross-Modality Learning.}}
To validate the efficacy of cross-modality learning, we turn off the prediction tasks, i.e., only keeping the reconstruction losses during training. Results have been shown in Fig.~\ref{fig:AE_gender}. We present the predicted functional networks of a randomly selected sample and the group average of the whole testing data. From the sparse structural networks, the corresponding functional connections have been correctly predicted and major patterns of the local network connections are captured. To further prove the accuracy, we conduct the statistical analysis on edges. Both direct and indirect edges in the target functional network are highly correlated with the predicted edges (Spearman correlation, overall is $r_S=0.83$ with $p<10^{-4}$), where the direct edges, $r_S=0.84$, are slightly greater than the indirect edges, $r_S=0.82$. We also prove the robustness of our model to the different sparsity levels of brain structural networks and results are shown in Supplementary Fig. 4.    



\vspace{-1em}
\subsection{Disease Classification}\label{sec:diseaseprediction}
\vspace{-0.5em}
In addition to the gender prediction in the healthy subjects, we retest our model on the disease classification. In this experiment, we include 323 subjects from Parkinson's Progression Markers Initiative (PPMI)~\cite{marek2011parkinson} and 224 of them are patients of Parkinson's disease (PD). We follow the experimental setting in gender prediction. $\mu_1=0.5$ and $\mu_2=0.5$ are used according to the grid search. 

\vspace{-1em}
\paragraph{\textbf{Classification Results.}} We consider the state-of-the-art baseline methods for comparison. The results are shown in Tab.~\ref{tab:genderprediction} (PPMI). Our model achieves the best prediction performance than other models (improving the accuracy by 5\% than BrainNetCNN, 9\% than Brain-Cheby and other baselines). Moreover, It also shows adding the cross-modality reconstruction do upgrade the performance. We locate the 10 key regions associating with the PD classification via the saliency map, see Supplementary Fig. 3. Most of the salient regions locate at the subcortical structures, such as the bilateral hippocampus and basal ganglia. These structures are conventionally conceived as the biomarkers of PD in medical imaging analysis~\cite{obeso2000pathophysiology,camicioli2003parkinson}.

\vspace{-1mm}
\section{Conclusion}
\vspace{-1mm}
We propose a novel multimodal brain network fusion framework based on a deep graph modal. The cross-modality network embedding is generated by an encoding-decoding network. The network embedding is also supervised by the prediction tasks. Eventually, the learned node features contribute to the brain saliency map for detecting disease-related biomarkers. In the future, we plan to extend our model to other learning tasks such as brain cortical parcellation and cognitive activity prediction.

\subsubsection*{Acknowledgments} This work was supported in part by NIH (RF1AG051710 and R01EB025032). We also gratefully acknowledge the support of NVIDIA Corporation with the donation of the Titan Xp GPU used for this research.

\bibliographystyle{splncs03}

\vspace{-1em}
\bibliography{BrainSurfaceNet}

\end{document}